\documentclass{article}
\usepackage[preprint, nonatbib]{neurips_2021}

\usepackage[utf8]{inputenc}
\usepackage[T1]{fontenc} 
\usepackage{amssymb,amsmath}
\usepackage{pifont}
\usepackage{graphicx,verbatimbox}
\usepackage{hyperref} 
\usepackage{tcolorbox}

\usepackage{xspace}
\usepackage{booktabs}
\usepackage{multirow}
\usepackage{rotating}
\usepackage{color}
\usepackage{palatino}

\newcommand{\cmark}{\ding{51}\xspace}%
\newcommand{\xmark}{\ding{55}\xspace}%
\def \alambic {\includegraphics[width=0.02\linewidth]{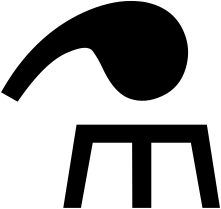}\xspace}

\title{Knowledge distillation to effectively attain both region-of-interest and global semantics from an image where multiple objects appear}
\author{
\begin{minipage}{\linewidth}
\begin{center}
\large Seonwhee Jin \\ 
\small \texttt{jin.seonwhee@gmail.com} \\
\end{center}
\end{minipage}
}

\date{~}

\begin{document}

\maketitle

\begin{abstract}
Models based on convolutional neural networks (CNN) and transformers have steadily been improved. They also have been applied in various computer vision downstream tasks. However, in object detection tasks, accurately localizing and classifying almost infinite categories of foods in images remains challenging. To address these problems, we first segmented the food as the region-of-interest (ROI) by using the segment-anything model (SAM) and masked the rest of the region except ROI as black pixels. This process simplified the problems into a single classification for which annotation and training were much simpler than object detection. The images in which only the ROI was preserved were fed as inputs to fine-tune various off-the-shelf models that encoded their own inductive biases. Among them, Data-efficient image Transformers (DeiTs) had the best classification performance. Nonetheless, when foods’ shapes and textures were similar, the contextual features of the ROI-only images were not enough for accurate classification. Therefore, we introduced a novel type of combined architecture, RveRNet, which consisted of ROI, extra-ROI, and integration modules that allowed it to account for both the ROI’s and global contexts. The RveRNet’s F1 score was 10\% better than other individual models when classifying ambiguous food images. If the RveRNet’s modules were DeiT with the knowledge distillation from the CNN, performed the best. We investigated how architectures can be made robust against input noise caused by permutation and translocation. The results indicated that there was a trade-off between how much the CNN teacher’s knowledge could be distilled to DeiT and DeiT’s innate strength. Code is publicly available at: \url{https://github.com/Seonwhee-Genome/RveRNet}.
\end{abstract}

\section{Introduction}
\label{sec:introduction}

Convolutional neural networks (CNN) have shown revolutionary performance in the ImageNet benchmark test~\cite{Deng2009, Russakovsky2015}and so have become de facto in computer vision use. Vision transformers (ViT), which apply self-attention~\cite{Vaswani2017} to analyzing images, do not suffer from CNNs’ inductive bias~\cite{Dosovitskiy2021}.

Photographed foods are typically shown on plates or, in the case of liquids, in glasses, cups, or bowls. Many images contain foods of various types and sizes. Different foods can be mixed in a dish (Figure 1(a)). Some foods, like water and colorless liquors, are almost impossible to distinguish from each other (Figure 1(b)). Therefore, it is challenging to localize and classify foods in images for applications like calorie measurement~\cite{Reddy2019}. Like with many natural image classifications, the criteria for categorizing foods are subjective, so certain categories can become too broad. Moreover, it is difficult to make finer-grained datasets. For example, although croissants and pretzels have completely different shapes, they can both be categorized as bread. There are almost infinite types of bread, so it is impossible to categorize breads by sub-types. 

The segment-anything model (SAM), which is the foundational model for promptable instance segmentation, can be prompted by coordinates and bounding boxes that specify the location of the instance~\cite{Kirillov2023}. This model makes previously impossible downstream tasks possible. Our study exploits the SAM’s promptable and fine instance segmentation ability to localize regions of interest (ROI) to simplify localization and classification problems to classification problems.

\begin{figure}[t]
    \centering
    \includegraphics[width=0.72\linewidth]{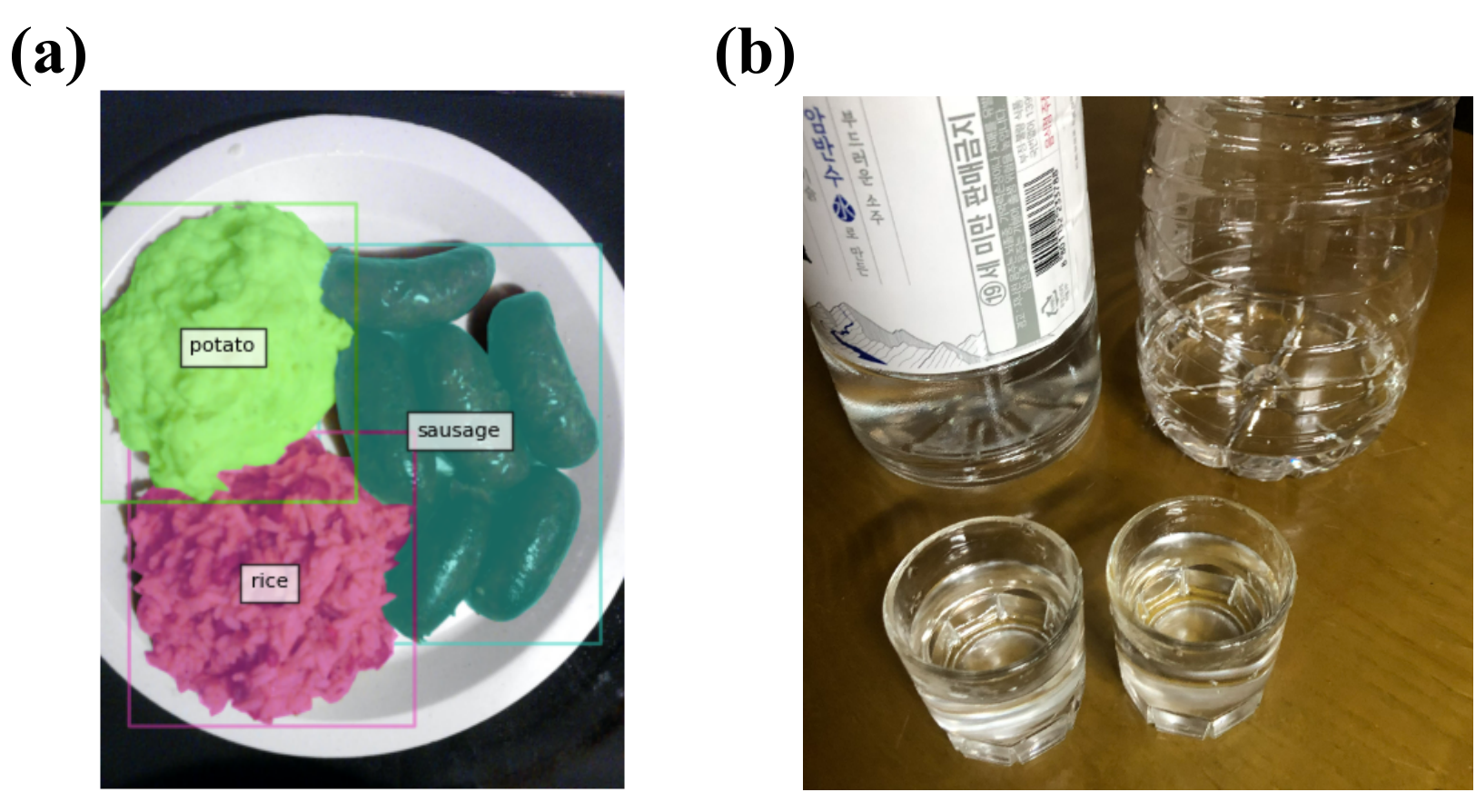}
    \caption{\textbf{Natural food images.} (a) Multiple kinds of foods on one plate. Annotated segmentation masks are shown. The original image is from the FoodSeg103 dataset. (b) Without any contextual clues, it is almost impossible to accurately classify which one is water and liquor. Photo by Seonwhee Jin, 2024. 
    \label{fig:fig1}}
\end{figure}

In this study, we investigated whether our proposed model can successfully overcome  the challenges with food datasets described above. We exploited the SAM’s instance segmentation ability to perform instance localization with high fidelity to classify segmented objects to reduce annotation costs and perform finer-grained classification more efficiently.

Our proposed ROI-vs-extra-ROI network (RveRNet) enabled us to properly exploit inductive biases from CNN and transformer model architectures. RveRNet successfully classified ambiguous food by taking global image context into consideration via the extra-ROI module.  %

Our results and contributions can be summarized as follows:
\begin{itemize}
    \item We simplified complex multiple object detection tasks as classification tasks by using SAM’s promptable instance segmentation. %
    \item We used RveRNet architecture to consider an image’s ROI via an ROI module and its global context via an extra-ROI module to improve ambiguous food classification. 
    \item 
    RveRNet performed best with DeiT$\alambic$ modules. This result may be explained by the combination of inductive biases of the CNN and transformers. However, the combination via knowledge distillation in DeiT might have caused a trade-off the CNN’s inductive bias and DeiT’s innate strength.     
\end{itemize}

\section{Related work}
\label{sec:related}

\subsection{Inductive Bias in Classification}
Inductive bias is the assumption that learning algorithms have their own properties that allow them to achieve their goals. For example, despite convolution tending to cause translation equivariance, CNNs that combine the convolution with other layers, like pooling, tend to cause translation invariance~\cite{Goyal2022, Biscione2021}. Thus, encoded inductive biases can be changed by designing learning algorithms’ architectures. For instance, locality can be assumed to be an inductive bias encoded in a CNN where input pixels spatially closer to each other are thought to be more correlated. In the context of image inputs, this assumption is qualitatively valid because local patches usually have closely related colors, textures, and lighting~\cite{Brendel2019, Naseer2021}. Multichanneling and downsampling CNNs contribute to capturing long-range correlation~\cite{Wang2023}.

\begin{table}[t]
\centering
\scalebox{0.9}
{
\begin{tabular}{lccc}
\toprule
Methods & ViT-B  & DeiT-B \\
\midrule
Epochs   & 300 & 300     \\
\midrule
Batch size & 4096 & 1024\\
Optimizer & AdamW & AdamW\\
     learning rate       & 0.003 &   $0.0005\times \frac{\textrm{batchsize}}{512} $  \\
     Learning rate decay & cosine & cosine  \\
     Weight decay        & 0.3    & 0.05    \\
     Warmup epochs  & 3.4 & 5       \\
\midrule
     Label smoothing $\varepsilon$ & \xmark & 0.1     \\
     Dropout      & 0.1 & \xmark     \\
     Stoch. Depth & \xmark & 0.1 \\
     Repeated Aug & \xmark & \cmark \\
     Gradient Clip. & \cmark & \xmark \\
\midrule
     Rand Augment  & \xmark        & 9/0.5 \\
     Mixup prob.  & \xmark & 0.8     \\
     Cutmix prob.   & \xmark & 1.0    \\
     Erasing prob.    & \xmark & 0.25    \\
 \bottomrule
\end{tabular}}
\caption{
 Ingredients and hyper-parameters for DeiT-B and ViT-B models (reprinted from~\cite{Touvron2021}).
\label{tab:comp_hyperparameters}}
\end{table}

Vision transformers (ViT) that analyze image patch sequences and lack the inductive biases of CNNs can classify images successfully~\cite{Dosovitskiy2021}. Since ViTs were created, studies comparing the inductive biases of CNNs and transformers have been conducted (e.g.,~\cite{Naseer2021, Raghu2021}). Studies have also been conducted in which inductive biases strongly encoded in one model were encoded in another model using knowledge distillation (e.g.,~\cite{Ren2022, Touvron2021}) and in which models have parallel architectures~\cite{Xu2024}.

Our investigation examined the strengths and weakness of ViTs at generalizing because they lack the inductive biases of CNNs, such as locality and weight sharing, as shown by analyzing the FoodSeg103 dataset~\cite{Wu2021}, the fact that Data-efficient image Transformers (DeiT) have the same architectures as ViTs but different training strategies (Table 1,~\cite{Touvron2021}), and that DeiT$\alambic$s, which were influenced by  CNN teachers’ inductive biases, perform better than ViTs and CNNs. 

ViTs based on self-attention have inductive biases like permutation invariance~\cite{Naseer2021} and permutation equivariance~\cite{Xu2024}. These inductive biases are not correlated to positional encoding~\cite{Naseer2021, Xu2024}.

\subsection{Promptable Segmentation that Enables ROI Separation}
SAMs are promptable instance segmentation models~\cite{Kirillov2023}, so they can be applied in a wide range of image types and analysis tasks. Prompts can be object coordinates, a bounding box enclosing an object, or a natural language description of the object. SAM variants, like FastSAM~\cite{Zhao2023} and FoodSAM~\cite{Lan2023}, exploit the original SAM’s robust segmentation performance and promptability. We used SAM to make our proposed RveRNet applicable to real-world classification problems.

\subsection{Consideration of Global Semantic Contexts}
By using SAM as our foundation model, identifying an ROI, and cutting out other regions~\cite{DeVries2017}, we were able to classify target foods in images in which there were multiple foods.  
However, in some cases, we needed to consider the global context because the ROI did not provide enough information. For example, it is difficult to distinguish between water and colorless liquors based only on the ROI (Figure 1(b)). Similarly, in South Korea, raw fish is commonly dipped in chili paste, while French fries are commonly dipped in ketchup. Due to their similar colors and textures, it is difficult to determine which sauce is in the image without considering the surrounding foods. Thus, it is sometimes helpful to consider both the ROI and the rest of the image because it may contain foods that are more likely to be paired with the food in the ROI. For example, the presence of French fries would indicate that the sauce in the ROI is ketchup, not chili paste.

\section{Methods} 
\label{sec:vit} 
\subsection{ROI Segmentation from the Raw Image}
With a given prompt, the SAM can finely segment the ROI , so we assumed that there were only ignorable discrepancies between real prompted segmentation results and the annotations of the FoodSeg103 dataset that we used to train and test our classifiers. Therefore, in a real-world application, raw input images could be processed using SAM or FastSAM.

\begin{figure}[t]
\centering \includegraphics[width=1\linewidth]{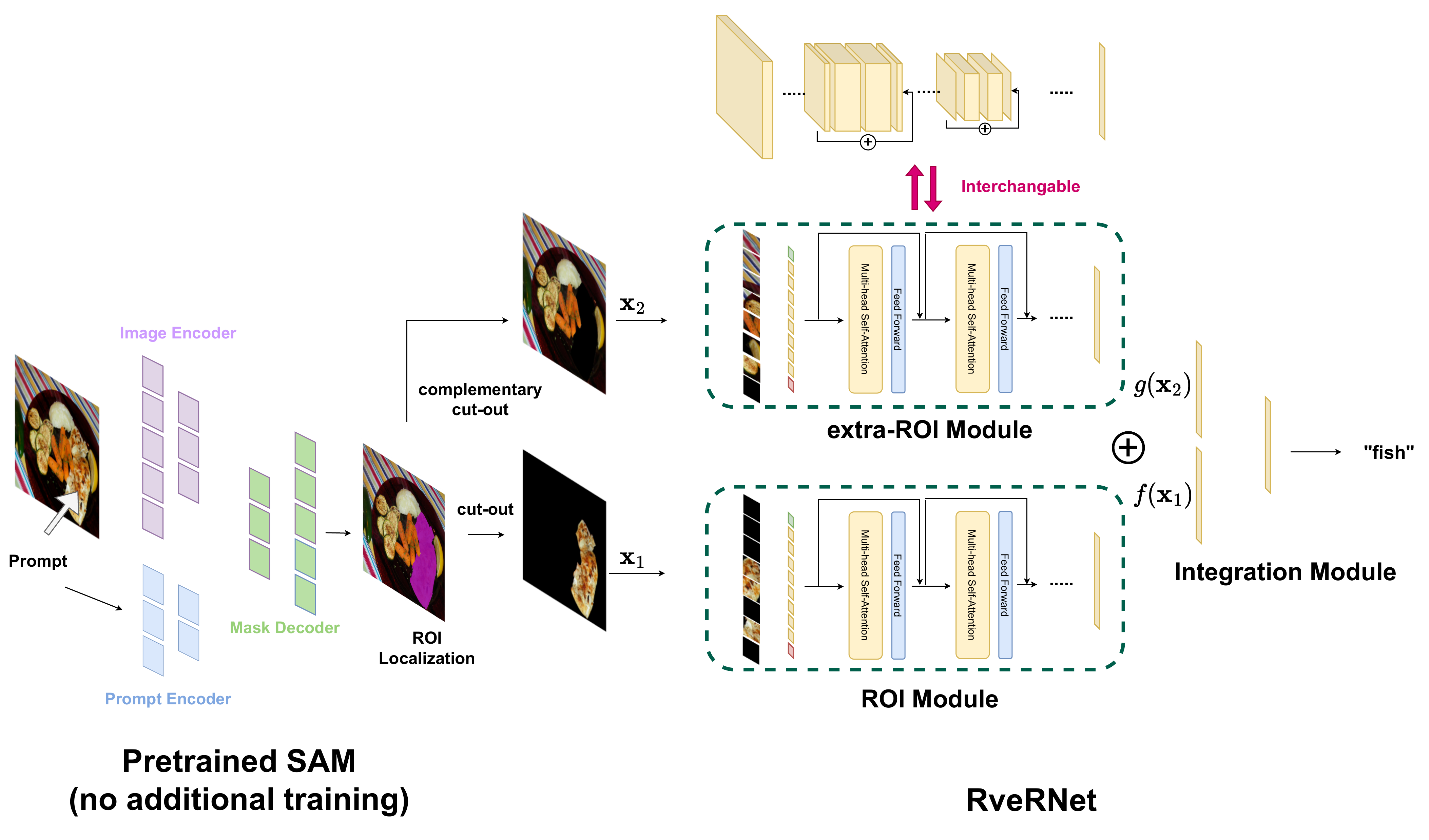}
\caption{\textbf{The structure of the proposed RveRNet} 
We used the robust SAM foundation model to segment the ROI of input images. Then, we processed the images to produce complementary cut-out pairs that were used as inputs for both the ROI and extra-ROI modules. The ROI and extra-ROI modules can have different architectures that encode different inductive biases.
\label{fig:fig2}}
\end{figure}

\subsection{ROI-vs-extra-ROI Consideration}
As described in Section 2.3, some foods cannot be accurately identified based only on the ROI. For example, ketchup and chili paste have similar colors and textures so foods outside of the ROI, like French fries, meatballs, onion rings, sashimi, and parboiled octopus, can be used to help distinguish between them.

To effectively consider both ROIs and extra-ROI regions, we designed a Siamese-like architecture consisting of ROI, extra-ROI , and integration modules. The ROI and extra-ROI modules did not necessarily have Siamese networks and so were allowed to have heterogeneous models.  The proposed RveRNet model was as follows:

\begin{equation}
p_\theta(\mathbf{x}_1,\mathbf{x}_2) =\textup{Softmax}(\textup{Linear}(\textup{ReLU}(\textup{Linear}(\textup{Concat}(f(\mathbf{x}_1),g(\mathbf{x}_2))))))
\end{equation}
\begin{equation}
= h((f(\mathbf{x}_1),g(\mathbf{x}_2)))
\end{equation}

Even for images containing multiple foods, ROIs were localized by SAM instance segmentation. The input for the ROI module , denoted as $\mathbf{x}_1$ in Equation 1 and 2, was the region outside of the ROI, which was masked by giving pixels in it color values of $0$, turning them black and allowing the ROI module to concentrate on the ROI. The input for the extra-ROI module, denoted as $\mathbf{x}_2$ in Equation 1 and 2, was the complementary cut-out produced by the ROI module in which the ROI pixels were masked in black while the extra-ROI region was preserved. The extra-ROI module, denoted as function $g$, output the global context whereas the ROI module, denoted as function $f$, output the crucial features used to identify the ROI.

The receptive fields of both the ROI and extra-ROI modules were concatenated, and then the concatenated product, containing information obtained from both modules, was processed in additional linear layers. This integrated module is denoted as function $h$.

\section{Experiments}
\label{sec:experiments}

\subsection{Dataset and Implementation Details}
We trained and evaluated our proposed RveRNet on the preprocessed FoodSeg103 dataset. To quantify the advantage of our proposed model’s architectures’ unique inductive biases, we avoided selecting a dataset that was too large for fine-tuning. Preprocessing the selected dataset created complementary cut-out images that masked the ROIs and were input into the extra-ROI module while the ROI images were input into the ROI module. 

In addition, to determine the degree to which the extra-ROI module in RveRNet enhanced the classification performance of ambiguous foods, we added images of ketchup and chili paste either photographed or collected from the internet, to the dataset. There were 69 ketchup and 72 chili paste train images and 38 ketchup and 34 chili paste test images. Thus, 18,320 train images and 7,769 test images across 105 categories were used in this study.

Unless otherwise specified, the train image dimensions were $224\times 224$ pixels . The off-the-shelf MobileNetV2, ViT, DeiT, DeiT$\alambic$, and MLP-Mixer models (~\cite{Dosovitskiy2021, Sandler2018, Touvron2021, Tolstikhin2021}) were pre-trained on ImageNet-1K for each RveRNet module. We optimized the proposed model’s parameters using the Adam optimizer~\cite{Kingma2015}, a linear warm-up, and a cosine learning rate decay scheduler. We applied a random horizontal flip to a train set with a 50\% probability. The training batches contained 50 images and had an initial learning rate of $4\times 10^{-3}$. Top-1 accuracy and F1 scores on the test set were calculated for models, which were fine-tuned over 30 epochs using one NVIDIA A100 GPU in Google Colaboratory Pro~\cite{Bisong2019}. 
Test results are presented in Table 2. The models were implemented using PyTorch~\cite{Paszke2019}, torchvision~\cite{maintainers2016}, and TIMM~\cite{Wightman2019}.

\subsection{Object Identification Based on Spatial ROI}

\begin{table}[t]
\caption{\textbf{The classifiers’ F1 scores for the test set of FoodSeg103 and ketchup and chili paste images.}
In all tables, the names and corresponding metrics of the best- and worst-performing models are bolded and underlined, respectively.  
\label{tab:tab2}}
\smallskip
\centering
\scalebox{0.8}
{
\begin{tabular}{l|cccccccc}
\toprule
\multirow{2}{*}{Model}       & F1 score          & F1 score   \\
            & (The whole dataset) & (Only for ketchup and chili paste) \\
\midrule
MobileNetV2   &  0.3765  &  0.5416 \\
ViT-B/16  &  0.3613  &  0.5085 \\
Mixer-B/16   &  0.3638   & 0.6167 \\
DeiT-B   &  0.3893   & 0.4873 \\
\textbf{DeiT-B$\alambic$}   &  \textbf{0.4289}   & \textbf{0.6238} \\
\bottomrule
\end{tabular}}
\end{table}

Classifiers extracted the necessary and important features from the extra-ROI images based on the images’ shapes, textures, and spatial information. The dataset’s variance was higher than that of other natural category datasets because the images were taken in various locations with various platings, decorations, and cooking status. Therefore, a certain architecture’s inductive bias may cause it to not classify food accurately.

\begin{figure}[t]
    \centering
    \includegraphics[width=1\linewidth]{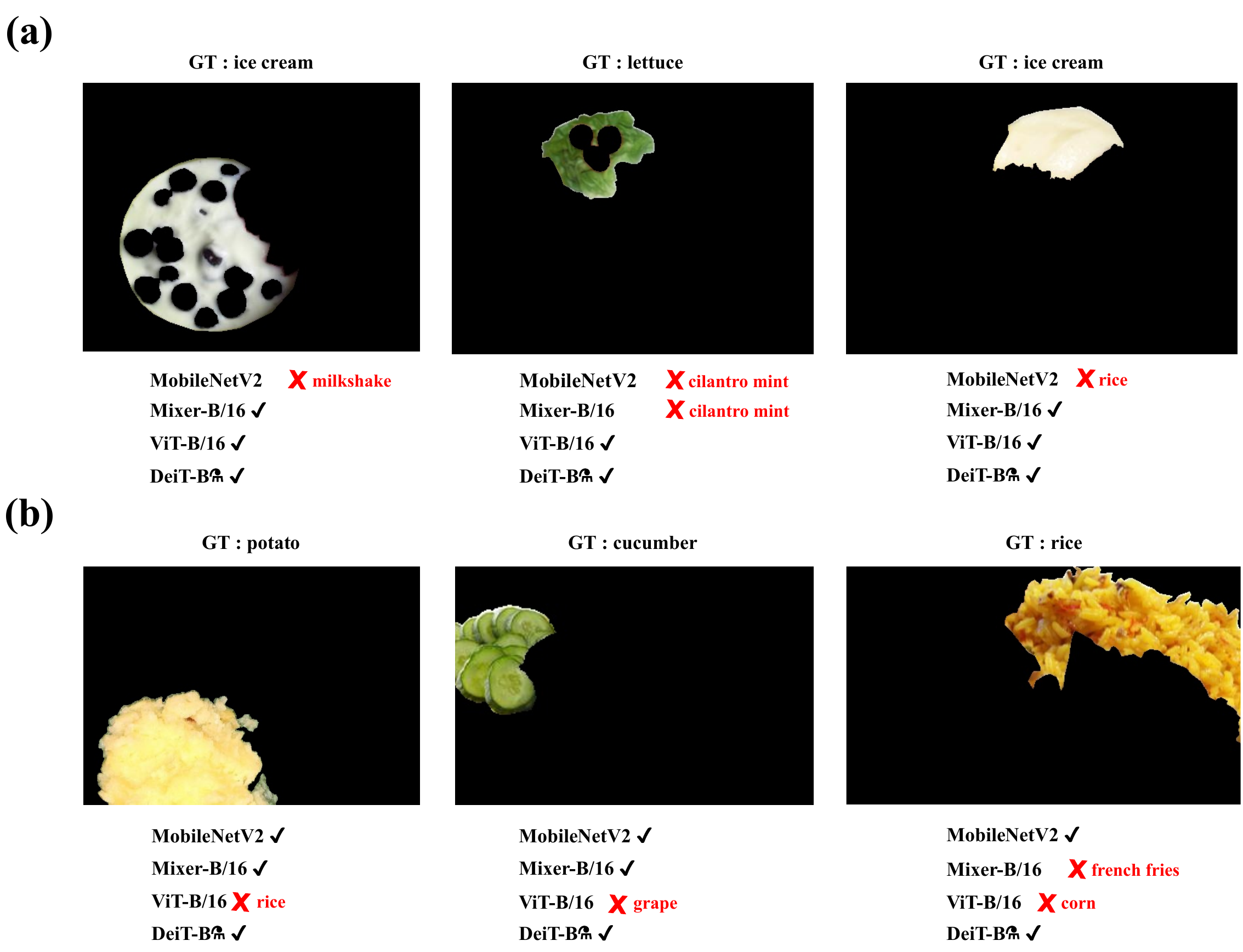}
    \caption{\textbf{Examples of classification by various off-the-shelf models.} (a) A case in which transformer-based models succeeded and the CNN failed. (b) A case in which the CNN model succeeded and the transformer-based models failed. In all figures and tables, we abbreviated “ground truth” as “GT” and “prediction as “pred.”  
    \label{fig:fig3}}
\end{figure}

For example, when foods were arranged irregularly, fragmented to the point that they lost a distinctive shape, or were randomly shaped, like ice cream, MobileNetV2 failed more than vision transformers with ViT by correctly classifying 530 of 7,769 images (Figure 3(a)).
MobileNetV2(CNN) is robust when the texture of the food is clear~\cite{Cohen2017} and in this study the CNN correctly classified 642 of 7,769 images (Figure 3(b)).  DeiT$\alambic$, which is based on ViT architecture but with the knowledge of a CNN because it can be distilled from a CNN teacher, performed the best in terms of F1 score and robustness (Figure 3).

\subsection{Classifications of Easily Identifiable and Ambiguous Foods}
\label{sec:distillation_results}

To identify the food in ROIs, the extra-ROI module captured  the global context that was then integrated with the features identified by the ROI module. When classifying ketchup and chili paste, RveRNet performed the best, which indicates that considering the global context via the extra-ROI module. RveRNet increased performance when the food has ambiguous shape and texture.

In order to achieve food classification using off-the-shelf models for the ROI and extra-ROI modules, we first must assume that the global context can be obtained by observing the entire image. The global context is the extra-ROI area, so CNNs that have locality inductive biases may be disadvantageous while self-attention-based ViTs may be more suitable because they can combine distant patches.

RveRNet performed better at classifying ketchup and chili paste when both the ROI and extra-ROI modules were MobileNetV2 (CNNs) than when only the ROI module was. However, it performed worse for the overall test set than the off-the-shelf models, which indicates that using CNNs’ inductive biases for both modules is ineffective. Furthermore, when both modules were either ViT or DeiT (self-attention-based models), the proposed model’s F1 scores, including for ketchup and chili paste, increased significantly (Table 3).

\begin{figure}[t]
    \centering
    \includegraphics[width=1\linewidth]{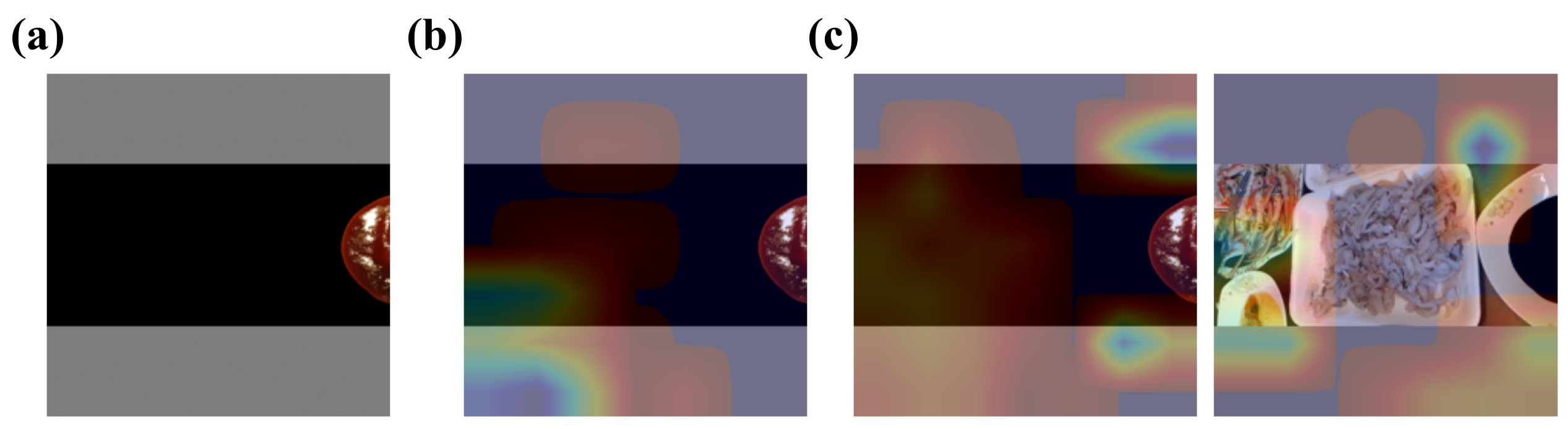}
    \caption{\textbf{GradCAM visualization chili paste images.} (a) Original ROI image. (b) Prediction by individual MobileNetV2 network. (c) Prediction by RveRNet with both modules being MobileNetV2.  
    \label{fig:fig4}}
\end{figure}

According to the GradCAM~\cite{Selvaraju2020} visualization, off-the-shelf models and the RveRNet model extracted different patterns and selected different information because contextual features and information identified by the extra-ROI module affect the forward step and error backpropagation during training RveRNet (Figure 4). The results for MobileNetV2 alone and RveRNet with both modules being MobileNetV2 can be explained by the assumption that feature extraction by the extra-ROI module affected classification performance both positively and negatively. Considering the extra-ROI module’s identified information can increase classification F1 scores for ambiguous foods (Table 3). However, it can also reduce ROI module optimization efficacy.

\begin{table}[t]
\caption{F1 scores of models for test sets of FoodSeg103 and ketchup and chili paste images combined and ketchup and chili paste images separately.  
\label{tab:tab3}}
\smallskip
\centering
\scalebox{0.8}
{
\begin{tabular}{l|cccccccc}
\toprule
\multirow{2}{*}{Model}  & \multirow{2}{*}{F1 score}  & F1 score   \\
 & & (only for ketchup and chili paste) \\
\midrule
MobileNetV2 & 0.3765 & 0.5416 \\
DeiT-B$\alambic$ & 0.4289 & 0.6238 \\
\midrule
RveRNet (ROI: MobileNetV2, extra-ROI: MobileNetV2) & 0.3347 & 0.6847 \\
RveRNet (ROI: MobileNetV2, extra-ROI: ViT-B/16) & 0.3722 & 0.5426 \\
RveRNet (ROI: ViT-B/16, extra-ROI: ViT-B/16) & 0.3821 & 0.7035 \\
RveRNet (ROI: ViT-B/16, extra-ROI: MobileNetV2) & 0.371 & 0.7905 \\
RveRNet (ROI: MobileNetV2, extra-ROI: DeiT-B$\alambic$) & 0.3755 & 0.7385 \\
RveRNet (ROI: DeiT-B, extra-ROI: DeiT-B) & 0.4155 & 0.7468 \\
\textbf{RveRNet (ROI: DeiT-B$\alambic$, extra-ROI: DeiT-B$\alambic$)} & \textbf{0.4424} & \textbf{0.8286} \\
\bottomrule
\end{tabular}}
\end{table}

When the ROI module is MobileNetV2 and the extra-ROI module is either ViT or DeiT, the F1 score for the whole test set was comparable to that of an individual MobileNetV2 model while the F1 score for the ketchup and chili paste alone increased (Table 3).
When ViT was the ROI module, regardless of the extra-ROI module’s architecture, RveRNet’s overall F1 score was much higher than when an individual ViT model trained the ROI region. When the extra-ROI module was MobileNetV2, the proposed model’s classification accuracy of ketchup and chili paste increased significantly (Table 3). This result is a product of CNNs’ ability to capture distant correlations~\cite{Wang2023}. Therefore, RveRNet's ROI module should have a transformer-based model, such as ViT or DeiT.

\begin{figure}[t]
    \centering
    \includegraphics[width=1.2\linewidth]{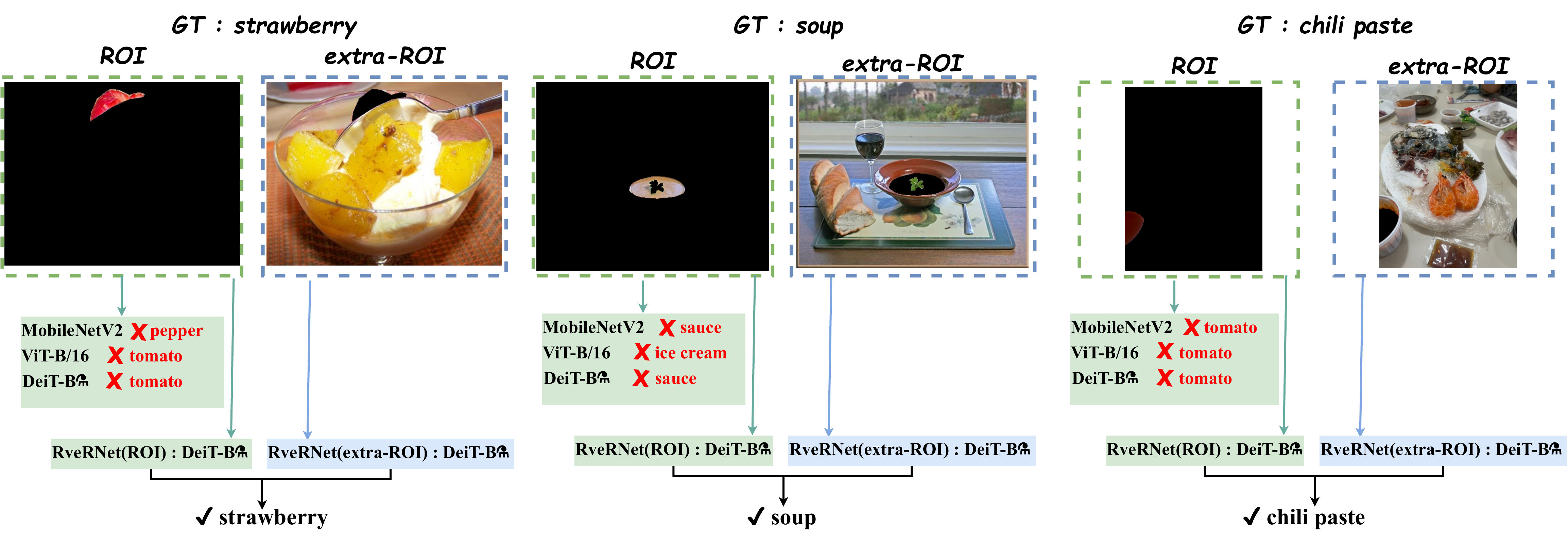}
    \caption{\textbf{Cases where individual off-the-shelf models failed but RveRNet succeeded.} The parallel structure of RveRNet enables it to take an image’s global context into account to more accurately classify ambiguous foods like ketchup and chili paste.
    \label{fig:fig5}}
\end{figure}

When both modules of RveRNet were DeiT$\alambic$, its classification performance was better than any RveRNet with off-the-shelf models RveRNet and any individual model (Table 2). This result indicates that transformer-based models that retain CNNs’ inductive bias obtain features from the ROI and global context well. 

The FoodSeg103 dataset includes a sauce category so, in many cases, ketchup and chili paste were misclassified as sauces (Figure 5). Therefore, we observed how classification performance varied for each model architecture when sauce-category images were removed from the train and test datasets (Table 8, Supplementary A). As a result, when RveRNet had DeiT$\alambic$ for both modules, it still maintained the best classification performance (Table 9, Supplementary A).

\subsection{Ablation Study}

Ketchup and chili paste are red dipping sauces with standardized container forms and low variation in their surrounding contexts. Foods that typically come with ketchup are chips, onion rings, and sausages, so they could be accurately classified without an ROI module.

\begin{figure}[t]
    \centering
    \includegraphics[width=1\linewidth]{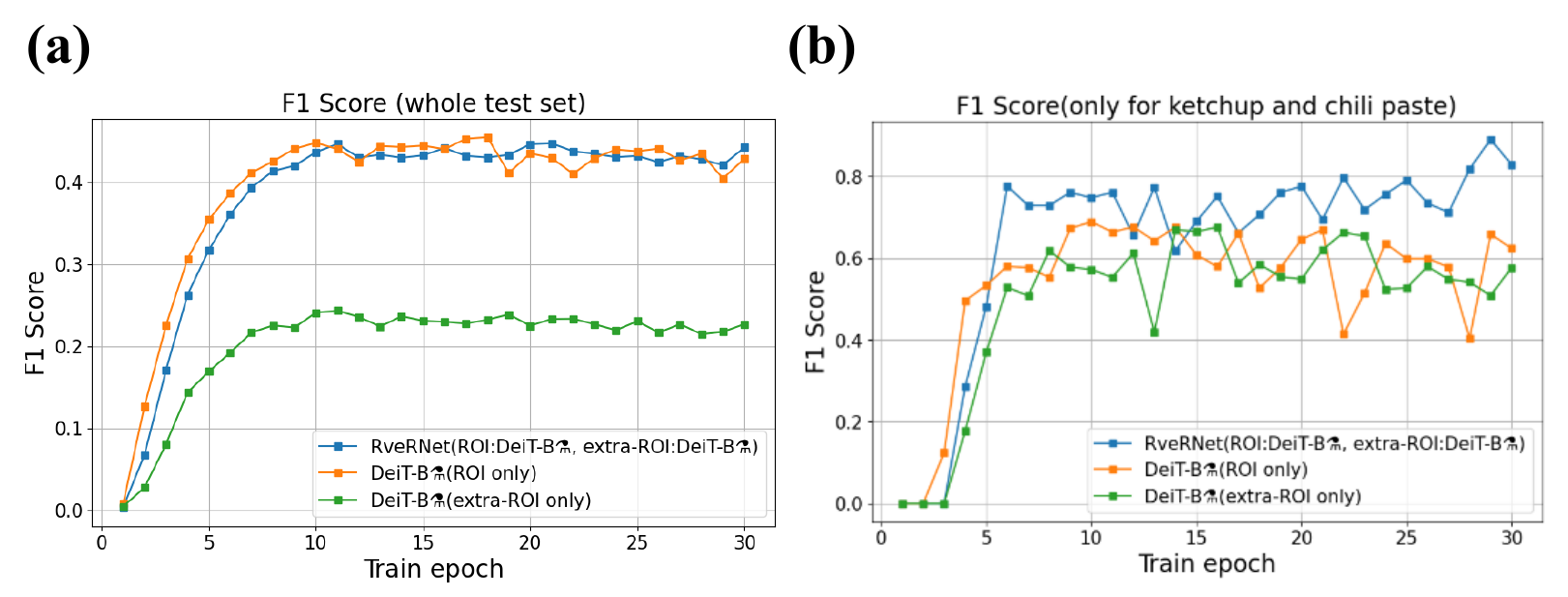}
    \caption{\textbf{F1 scores at each training step.} RveRNet with DeiT-B$\alambic$ as both modules performed the best in general and for ketchup and chili paste.
    \label{fig:fig6}}
\end{figure}

\begin{table}[t]
\caption{\textbf{Ablation study results for RveRNet.} The F1 score for the overall test set without the ROI module was significantly lower than when the ROI module was included. F1 scores indicated that RveRNet accurately classified ketchup and chili paste.  
\label{tab:tab4}}
\smallskip
\centering
\scalebox{0.8}
{
\begin{tabular}{l|cccccccc}
\toprule
\multirow{2}{*}{Model} & \multirow{2}{*}{F1 score}  & F1 score   \\
 & & (only for ketchup and chili paste) \\
\midrule
MobileNetV2 (ROI only) &  0.3765 &  0.5416 \\
ViT-B/16 (ROI only) & 0.3613 &  0.5085 \\
DeiT-B$\alambic$ (ROI only  &  0.4289 & 0.6238 \\
\midrule
MobileNetV2 (extra-ROI only)  &  0.1969  &  \underline{0.4573} \\
ViT-B/16 (extra-ROI only) &  \underline{0.1934} &  0.5625 \\
DeiT-B$\alambic$ (extra-ROI only) &  0.2269  & 0.5766 \\
\midrule
RveRNet (ROI: MobileNetV2, extra-ROI: MobileNetV2)   &  0.3347  &  0.6847 \\
RveRNet (ROI: ViT-B/16, extra-ROI: ViT-B/16)   &  0.3821  &  0.7035 \\
\textbf{RveRNet (ROI: DeiT-B$\alambic$, extra-ROI: DeiT-B$\alambic$)}   &  \textbf{0.4424}  &  \textbf{0.8286} \\
\bottomrule
\end{tabular}}
\end{table}

We can replace  the results for RveRNet without the extra-ROI module with the results of the individual model architectures in Section 4.2. RveRNet performed significantly worse without an ROI module than with one RveRNet (Table 4, Figure 6). This result indicates that the ROI module is the main driver of performance but the extra-ROI module plays a crucial role when classifying ambiguous foods, namely ketchup and chili paste . Moreover, the parallel structure of the ROI and extra-ROI modules and the module that integrates their outputs in RveRNet allows it to analyze visual and contextual features from both the ROI and global context to correctly classify the food in ROI.

\begin{table}[t]
\caption{\textbf{The average top-1 accuracy declined when ROI module inputs were permuted.} 
}
\smallskip
\centering
\scalebox{0.88}
{
\begin{tabular}{l|c|c|cccc}
\toprule
 \multirow{4}{*}{ROI module} &  \multicolumn{3}{c}{Average top-1 accuracy declined} \\
 & \multicolumn{3}{c}{when the input was permuted in a 16$\times$16 patch} \\
\cline{2-4} 
 & ROI module & Extra-ROI module & Input permutations \\
 & input permutations & input permutations & for both modules \\
\midrule
 CNN (MobileNetV2) & \underline{-31.529} & \underline{-15.530} & \underline{-41.840} \\
 Transformer (ViT-B/16 and DeiT-B)  & \textbf{-16.804} & -8.927 & \textbf{-29.444}    \\ 
 DeiT-B$\alambic$  & -22.371   & \textbf{-6.822}  & -32.745 \\ 
\bottomrule
\end{tabular}}
\end{table}

\begin{table}[t]
\caption{\textbf{Average top-1 accuracy declined when extra-ROI module inputs were permuted.} 
}
\smallskip
\centering
\scalebox{0.88}
{
\begin{tabular}{l|c|c|cccc}
\toprule
\multirow{4}{*}{Extra-ROI module} &  \multicolumn{3}{c}{Average top-1 accuracy declined} \\
 &  \multicolumn{3}{c}{when the input was permuted in a 16$\times$16 patch} \\
 \cline{2-4} 
 & ROI module & Extra-ROI module & Input permutations \\
 & input permutations & input permutations & for both modules \\
\midrule
 CNN (MobileNetV2)  & \underline{-27.668} & \underline{-15.298}  & \underline{-38.899} \\
 Transformer (ViT-B/16 and DeiT-B)  & -20.505    & -8.899  &  -31.823    \\ 
 DeiT-B$\alambic$  & \textbf{-17.898}   & \textbf{-8.148}  & \textbf{-30.467} \\ 
\bottomrule
\end{tabular}}
\end{table}

\subsection{Inductive Bias of Classifiers that Affect Object Classification}

The input to the ROI module only preserves pixel values in the ROI and masks the extra-ROI in black . Therefore, to achieve a robust classification, it may be important for the ROI module’s model to encode translation invariance. However, the input of the extra-ROI module preserves every pixel value outside of the ROI so that CNNs’ inductive biases, like locality and translation equivariance, and transformers’ inductive biases, like permutation invariance, are necessary to obtain the global context robustly.
\begin{equation}
    h((f(\mathbf{x}_1),g(\mathbf{x}_2))) = h((f(\mathbf{P}\mathbf{x}_1),g(\mathbf{x}_2)))
\end{equation}
\begin{equation}
h((f(\mathbf{x}_1),g(\mathbf{x}_2))) = h((f(\mathbf{x}_1),g(\mathbf{P}\mathbf{x}_2))) 
\end{equation}
\begin{equation} 
h((f(\mathbf{x}_1),g(\mathbf{x}_2))) = h((f(\mathbf{P}\mathbf{x}_1),g(\mathbf{P}\mathbf{x}_2))) 
\end{equation}
We can express the permutation invariance in RveRNet as shown in equations 3–5. Equation 3 models the case when a permutation operation is applied to the input to the ROI module . Equation 4 models the case when the permutation operation is applied to the input to the extra-ROI module. Equation 5 models the case when the permutation operation is applied to both inputs to RveRNet. In Supplementary B.2, we compare the robustness of RveRNet based on the permutation invariance derived from the top-1 accuracy of each combination of modules in RveRNet (Figure 9).

We grouped off-the-shelf models by architecture type and calculated the average of each group’s top-1 accuracy decrease to summarize the permutation experiment’s results (Tables 5 and 6). When the ROI module was a CNN, RveRNet was vulnerable to input permutation noise. When the ROI module was DeiT$\alambic$, RveRNet’s performance decreased more than when it was DeiT. The CNN teacher’s distillation to DeiT may reduce the transformer’s inductive bias (Table 5). 

When the extra-ROI module was a CNN, RveRNet was vulnerable to input permutation noise. The transformers’ performance decline was comparable to that of DeiT-B$\alambic$ as extra-ROI modules (Table 6).

\begin{equation} 
h((f(\mathbf{x}_1),g(\mathbf{x}_2))) = h((f(\mathbf{x}_1),g(\mathbf{T}_a\mathbf{x}_2))) 
\end{equation}
\begin{equation} 
h((f(\mathbf{x}_1),g(\mathbf{x}_2))) = h((f(\mathbf{x}_1),g(\mathbf{T}_b\mathbf{x}_2))) 
\end{equation}
\begin{equation} 
h((f(\mathbf{x}_1),g(\mathbf{x}_2))) = h((f(\mathbf{x}_1),g(\mathbf{T}_c\mathbf{x}_2))) 
\end{equation}

We can express the translation invariance of RveRNet as shown in equations 6–8. Only the input to the extra-ROI module is translocated. In addition to information loss in the global context, the extra-ROI module’s input was translocated. In Supplementary B.3, we compare the robustness of RveRNet based on the translation invariance derived from the top-1 accuracy of each combination of modules in RveRNet (Figure 10).

\begin{table}[t]
\caption{\textbf{Average top-1 accuracy declined when ROI module inputs were translocated resolution-wise.} 
}
\smallskip
\centering
\scalebox{0.88}
{
\begin{tabular}{l|cccccccc}
\toprule
\multirow{3}{*}{Extra-ROI module} &  \multicolumn{3}{c}{Average top-1 accuracy declined} \\
& \multicolumn{3}{c}{when extra-ROI module inputs were translocated} \\ \cline{2-4}
& dx: -30, dy: 63 & dx: 60, dy: -85 & dx: 37, dy: 139 \\
\midrule
 CNN (MobileNetV2)  & -0.522  & -1.886  & -2.684 \\
 Transformer (ViT-B/16)   & \underline{-1.209}  & \underline{-5.412}  & \underline{-4.73} \\
 Transformer (DeiT-B)  & \textbf{1.956}   & \textbf{-0.644}  & \textbf{-0.799} \\
\midrule
 CNN’s knowledge distilled & \multirow{2}{*}{-0.863}  & \multirow{2}{*}{-4.28}  & \multirow{2}{*}{-2.922} \\
 Transformer(DeiT-B$)\alambic$  & & &  \\ 
\bottomrule
\end{tabular}}
\end{table}

We also grouped models by architecture type and determined their average top-1 accuracy drops to summarize the translocation experiment’s results. When the extra-ROI module was a CNN, RveRNet was relatively robust to information loss due to input translocation. This result may be explained by the translation invariance of CNNs. Although accuracy significantly declined when the extra-ROI module was ViT-B, it was not due the transformer-based models’ weak translation invariance because DeiT-B, which had the same architecture as ViT-B but different training strategies, distillation token, and the final layer for pre-training~\cite{Touvron2021}, showed the least performance drop, indicating its robustness (Table 7). However, when the extra-ROI module was DeiT-B$\alambic$, accuracy declined significantly. 

According to the permutation and translation invariance investigation results, we can conclude that there was a trade-off between CNNs’ knowledge distillation to DeiT-B and the transformer’s strength. Despite this trade-off, DeiT-B$\alambic$ still had good classification performance.

\section{Conclusion}
\label{sec:conclusion}

We attempted to cope with challenging problems of food localization and classification by exploiting SAM’s confident instance segmentation and our proposed RveRNet model that considered the global context of input images. Off-the-shelf models should be selected for RveRNet modules considering their inductive biases. RveRNet performed best when both ROI and extra-ROI modules were DeiT-B$\alambic$ due to their combination of knowledge from CNNs and transformers. Furthermore, we identified which inductive biases increased RveRNet’s robustness to noisy inputs. DeiT-B$\alambic$ performed worse than other architectures that encoded their own inductive biases but still performed meaningfully well. We can conclude that CNNs’ knowledge distillation to DeiT-B yielded a trade-off but did not affect the overall classification strength of DeiT-B$\alambic$. 

\subsection*{Acknowledgements}

Many thanks to Jonghee Lee for sharing his idea about the inductive bias of the MLP-mixer model and its implementation. We thank Haidan Liu and Hanna Seok for brainstorming and having valuable discussions with us. Thanks to Barrett Hunter, whose advice and comments improved this manuscript. We also thank to Jeonmo Kang and Ki-Eun Hyeong for taking ketchup and chili paste photos.

\clearpage

{
\bibliographystyle{ieeetr}
\bibliography{main}

\begin{thebibliography}{10}

\bibitem{Deng2009}
J.~Deng, W.~Dong, R.~Socher, L.-J. Li, K.~Li, and L.~Fei-Fei, ``Imagenet: A large-scale hierarchical image database,'' pp.~248--255, 2009.

\bibitem{Russakovsky2015}
O.~Russakovsky, J.~Deng, H.~Su, J.~Krause, S.~Satheesh, S.~Ma, Z.~Huang, A.~Karpathy, A.~Khosla, M.~Bernstein, A.~C. Berg, and L.~Fei-Fei, ``Imagenet large scale visual recognition challenge,'' {\em International Journal of Computer Vision (IJCV)}, vol.~115, pp.~211--252, 2015.

\bibitem{Vaswani2017}
A.~Vaswani, N.~Shazeer, N.~Parmar, J.~Uszkoreit, L.~Jones, A.~N. Gomez, Łukasz Kaiser, and I.~Polosukhin, ``Attention is all you need,'' vol.~30, Curran Associates, Inc., 2017.

\bibitem{Dosovitskiy2021}
A.~Dosovitskiy, L.~Beyer, A.~Kolesnikov, D.~Weissenborn, X.~Zhai, T.~Unterthiner, M.~Dehghani, M.~Minderer, G.~Heigold, S.~Gelly, J.~Uszkoreit, and N.~Houlsby, ``An image is worth 16x16 words: Transformers for image recognition at scale,'' {\em ICLR}, 2021.

\bibitem{Reddy2019}
V.~H. Reddy, S.~Kumari, V.~Muralidharan, K.~Gigoo, and B.~S. Thakare, ``Food recognition and calorie measurement using image processing and convolutional neural network,'' pp.~109--115, IEEE, 5 2019.

\bibitem{Kirillov2023}
A.~Kirillov, E.~Mintun, N.~Ravi, H.~Mao, C.~Rolland, L.~Gustafson, T.~Xiao, S.~Whitehead, A.~C. Berg, W.-Y. Lo, P.~Dollár, and R.~Girshick, ``Segment anything,'' {\em arXiv:2304.02643}, 2023.

\bibitem{Goyal2022}
A.~Goyal and Y.~Bengio, ``Inductive biases for deep learning of higher-level cognition,'' {\em Proceedings of the Royal Society A: Mathematical, Physical and Engineering Sciences}, vol.~478, 10 2022.

\bibitem{Biscione2021}
V.~Biscione and J.~S. Bowers, ``Convolutional neural networks are not invariant to translation, but they can learn to be,'' {\em Journal of Machine Learning Research}, vol.~22, pp.~1--28, 2021.

\bibitem{Brendel2019}
W.~Brendel and M.~Bethge, ``Approximating cnns with bag-of-local-features models works surprisingly well on imagenet,'' {\em International Conference on Learning Representations}, 2019.

\bibitem{Naseer2021}
M.~Naseer, K.~Ranasinghe, S.~Khan, M.~Hayat, F.~Khan, and M.-H. Yang, ``Intriguing properties of vision transformers,'' 2021.

\bibitem{Wang2023}
Z.~Wang and L.~Wu, ``Theoretical analysis of the inductive biases in deep convolutional networks,'' vol.~36, pp.~74289--74338, Curran Associates, Inc., 2023.

\bibitem{Touvron2021}
H.~Touvron, M.~Cord, M.~Douze, F.~Massa, A.~Sablayrolles, and H.~Jegou, ``Training data-efficient image transformers \& distillation through attention,'' vol.~139, pp.~10347--10357, 7 2021.

\bibitem{Raghu2021}
M.~Raghu, T.~Unterthiner, S.~Kornblith, C.~Zhang, and A.~Dosovitskiy, ``Do vision transformers see like convolutional neural networks?,'' 2021.

\bibitem{Ren2022}
S.~Ren, Z.~Gao, T.~Hua, Z.~Xue, Y.~Tian, S.~He, and H.~Zhao, ``Co-advise: Cross inductive bias distillation,'' 6 2022.

\bibitem{Xu2024}
H.~Xu, L.~Xiang, H.~Ye, D.~Yao, P.~Chu, and B.~Li, ``Permutation equivariance of transformers and its applications,'' 2024.

\bibitem{Wu2021}
X.~Wu, X.~Fu, Y.~Liu, E.-P. Lim, S.~C.~H. Hoi, and Q.~Sun, ``A large-scale benchmark for food image segmentation,'' 2021.

\bibitem{Zhao2023}
X.~Zhao, W.~Ding, Y.~An, Y.~Du, T.~Yu, M.~Li, M.~Tang, and J.~Wang, ``Fast segment anything,'' 2023.

\bibitem{Lan2023}
X.~Lan, J.~Lyu, H.~Jiang, K.~Dong, Z.~Niu, Y.~Zhang, and J.~Xue, ``Foodsam: Any food segmentation,'' {\em IEEE Transactions on Multimedia}, pp.~1--14, 2023.

\bibitem{DeVries2017}
T.~DeVries and G.~W. Taylor, ``Improved regularization of convolutional neural networks with cutout,'' {\em arXiv preprint arXiv:1708.04552}, 2017.

\bibitem{Sandler2018}
M.~Sandler, A.~Howard, M.~Zhu, A.~Zhmoginov, and L.-C. Chen, ``Mobilenetv2: Inverted residuals and linear bottlenecks,'' pp.~4510--4520, IEEE, 6 2018.

\bibitem{Tolstikhin2021}
I.~Tolstikhin, N.~Houlsby, A.~Kolesnikov, L.~Beyer, X.~Zhai, T.~Unterthiner, J.~Yung, A.~Steiner, D.~Keysers, J.~Uszkoreit, M.~Lucic, and A.~Dosovitskiy, ``Mlp-mixer: An all-mlp architecture for vision,'' {\em arXiv preprint arXiv:2105.01601}, 2021.

\bibitem{Kingma2015}
D.~Kingma and J.~Ba, ``Adam: A method for stochastic optimization,'' 2015.

\bibitem{Bisong2019}
E.~Bisong, {\em Google Colaboratory}, pp.~59--64.
\newblock Berkeley, CA: Apress, 2019.

\bibitem{Paszke2019}
A.~Paszke, S.~Gross, F.~Massa, A.~Lerer, J.~Bradbury, G.~Chanan, T.~Killeen, Z.~Lin, N.~Gimelshein, L.~Antiga, A.~Desmaison, A.~Kopf, E.~Yang, Z.~DeVito, M.~Raison, A.~Tejani, S.~Chilamkurthy, B.~Steiner, L.~Fang, J.~Bai, and S.~Chintala, {\em PyTorch: An Imperative Style, High-Performance Deep Learning Library}, pp.~8024--8035.
\newblock Curran Associates, Inc., 2019.

\bibitem{maintainers2016}
maintainers and contributors, ``Torchvision: Pytorch's computer vision library,'' 2016.

\bibitem{Wightman2019}
R.~Wightman, ``Pytorch image models,'' 2019.

\bibitem{Cohen2017}
N.~Cohen and A.~Shashua, ``Inductive bias of deep convolutional networks through pooling geometry,'' 2017.

\bibitem{Selvaraju2020}
R.~R. Selvaraju, M.~Cogswell, A.~Das, R.~Vedantam, D.~Parikh, and D.~Batra, ``Grad-cam: Visual explanations from deep networks via gradient-based localization,'' {\em International Journal of Computer Vision}, vol.~128, pp.~336--359, 2 2020.

\end{thebibliography}
}
\newpage
\appendix

\section*{Supplementary Materials}
\label{sec:supplement}

\begin{table}[t]
\caption{The number of models’ predictions when the sauce class was included (left) and excluded (right). \smallskip  
\label{tab:tab8}}
\centering
\scalebox{0.86}
{
\begin{tabular}{l|cc|cc}
\toprule 
            &  \multicolumn{2}{c}{GT : ketchup (\# GT : 38)}  & \multicolumn{2}{c}{GT : chili paste (\# GT : 34)} \\ \cline{2-5}
            & Pred: ketchup \cmark & Pred: chili paste \xmark &  Pred: chili paste \cmark  &  Pred: ketchup \xmark \\
\midrule 
MobileNetV2          & 22/24 (-2)   & 8/2 (+6)    & 22/14 (+8)  &  3/5 (-2)  \\
ViT-B/16    & 21/25 (-4)   & 6/2 (+4)    & 21/18 (+3)  & 4/3 (+1)   \\
DeiT-B$\alambic$          & 28/22 (+6)   & 5/0 (+5)    & 20/17 (+3)  &  6/4 (+2)  \\
\midrule 
RveRNet & \multirow{3}{*}{35/28 (+7)}   & \multirow{3}{*}{1/0 (+1)}    & \multirow{3}{*}{22/24 (-2)}  & \multirow{3}{*}{9/5 (+4)}   \\
(ROI: DeiT-B$\alambic$,    &  & &  &   \\
extra-ROI: DeiT-B$\alambic$)    &  & &  &   \\
\bottomrule
\end{tabular}}
\end{table}

\begin{table}[t]
\caption{F1 scores of models for test sets of FoodSeg103 and ketchup and chili paste images combined and ketchup and chili paste images separately.  
\label{tab:tab9}}
\smallskip
\centering
\scalebox{0.8}
{
\begin{tabular}{l|cc|cc}
\toprule
    & \multicolumn{2}{c}{F1 score when sauce class included} & \multicolumn{2}{c}{F1 Score when sauce class excluded} \\ \cline{2-5}
    & \multirow{2}{*}{whole dataset} & only for & \multirow{2}{*}{whole dataset} & only for \\ 
    & & ketchup and chili paste &  & ketchup and chili paste \\
\midrule
MobileNetV2 & 0.3765 & 0.5416 & 0.4084 & 0.6519 \\
ViT-B/16 &  0.3613 & 0.5085 & 0.3499 & 0.6615 \\
DeiT-B$\alambic$ & 0.4289 & 0.6238 & 0.4326 & 0.6715 \\
\midrule 
\textbf{RveRNet} & \multirow{3}{*}{\textbf{0.4424}} & \multirow{3}{*}{\textbf{0.8286}} & \multirow{3}{*}{\textbf{0.4366}} & \multirow{3}{*}{\textbf{0.7544}} \\
\textbf{(ROI: DeiT-B$\alambic$,} & & & & \\
\textbf{extra-ROI: DeiT-B$\alambic$)} & & & &  \\

\bottomrule
\end{tabular}
}
\end{table}

\section{Model Performance for Ambiguous Foods}

When the sauce class was excluded from the dataset, models’ ability to accurately classify ketchup and chili paste decreased (Table 8).

The results in Table 9 indicate that the global context analyzed by the extra-ROI module enhanced model performance when classifying ambiguous ketchup and chili paste images. When RveRNet had DeiT-B$\alambic$ for both modules, it did not perform worse than other single classifier architectures, which was not the case for other combinations of modules in RveRNet.

MobileNetV2’s F1 score increased after excluding the sauce category from the test set. Classification accuracy of ambiguous foods also increased. Although ketchup and chili paste looked similar and were distinctive from other sauces, images of these sauces had similar-looking ROIs that may have confused the classifiers.

\section{Data Augmentation to Investigate the Effect of Inductive Bias}
\subsection{Individual Off-the-shelf Models}

\begin{figure}[t]
    \centering
    \includegraphics[width=1\linewidth]{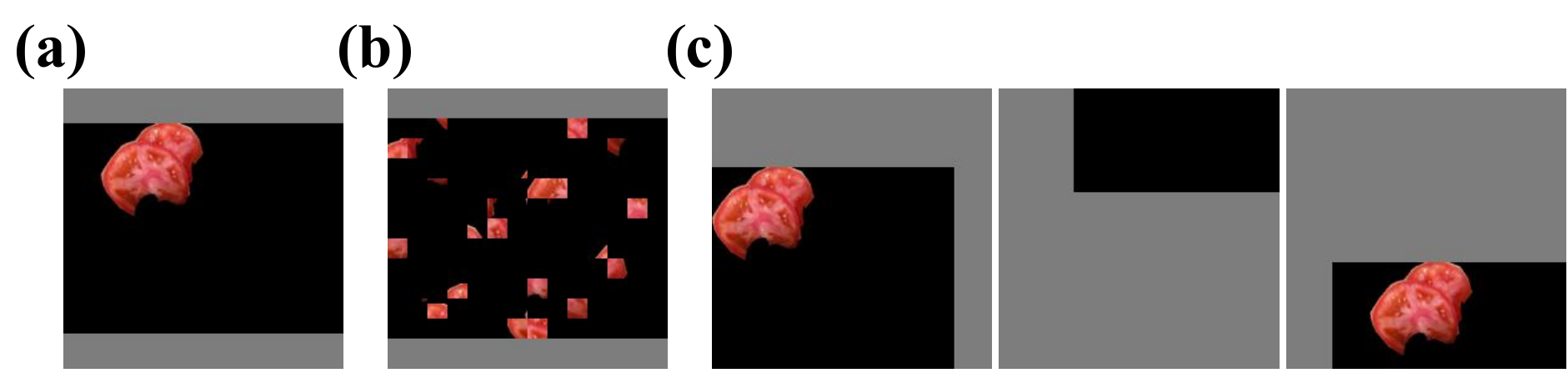}
    \caption{\textbf{Sample images from the FoodSeg103 dataset being instance-segmented and preprocessed.}(a) No transformation. (b) Permutation. (C) Translocation by (dx: -30, dy: 63), (dx: 60, dy: -85), and (dx: 37, dy: 139), respectively.
    \label{fig:fig8}}
\end{figure}

ROI translocation risks losing all of the information in the ROI (Figure 7(c)). Therefore, it is inappropriate to quantify translation invariance using randomly translocated images.

\begin{table}[t]
\caption{\textbf{Top-1 accuracy of each model and its decline when the input image was randomly permuted.}
\label{tab:tab10}}
\smallskip
\centering
\scalebox{0.8}
{
\begin{tabular}{l|cccccccc}
\toprule
    &  \multicolumn{5}{c}{Top-1 accuracy (test set)} \\ 
 & Mixer-B/16   & \textbf{ViT-B/16}  & \underline{MobileNetV2} & DeiT-B & DeiT-B$\alambic$   \\
\midrule
No transformation   & 54.370   & 51.757 & 54.833 & 55.902 & 59.248 \\
ROI permutation & 27.816   & 26.117 & 13.670 & 29.335 & 26.992\\
\midrule
Performance decline & \multirow{2}{*}{-26.554} & \multirow{2}{*}{\textbf{-25.640}} & \multirow{2}{*}{\underline{-41.164}} & \multirow{2}{*}{-26.567} & \multirow{2}{*}{-32.256} \\
due to ROI permutation & & & & & \\
\bottomrule
\end{tabular}}
\end{table}

The transformer-based models, such as ViT and DeiT, were robust against permuted inputs, which indicates that these models encoded relatively strong permutation invariance (Table 10). However, CNNs, like MobileNetV2, and CNNs’ knowledge-distilled transformers, like DeiT-B$\alambic$, were relatively vulnerable to input permutation. 

\subsection{Permutation Invariance}

\begin{table}[t]
\caption{\textbf{Top-1 accuracy when the inputs of RveRNet were permuted.}
\label{tab:tab11}}
\smallskip
\centering
\scalebox{0.8}
{
\begin{tabular}{l|cccccccc}
\toprule
    &  \multicolumn{4}{c}{Top-1 accuracy (test set)} \\ \cline{2-5}
 & No data   & \multicolumn{3}{c}{Random patch (16$\times$16) shuffling conducted for} \\ 
& transformation   & ROI module inputs  & extra-ROI module inputs & both modules’ inputs   \\ 
\midrule
ROI   &  \includegraphics[width=0.25\textwidth]{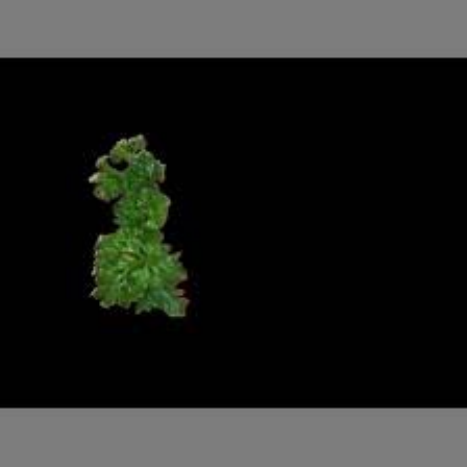}  & \includegraphics[width=0.25\textwidth]{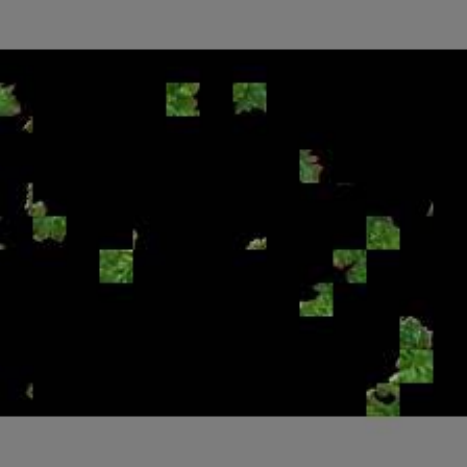}  & \includegraphics[width=0.25\textwidth]{figs/Figure9-2.pdf}& \includegraphics[width=0.25\textwidth]{figs/Figure9-1.pdf} \\
\midrule
Extra-ROI & \includegraphics[width=0.25\textwidth]{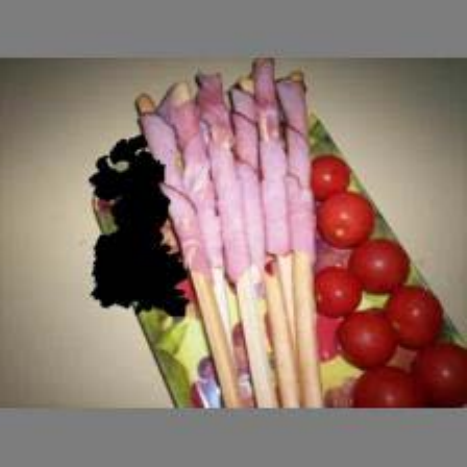}   & \includegraphics[width=0.25\textwidth]{figs/Figure9-3.pdf} & \includegraphics[width=0.25\textwidth]{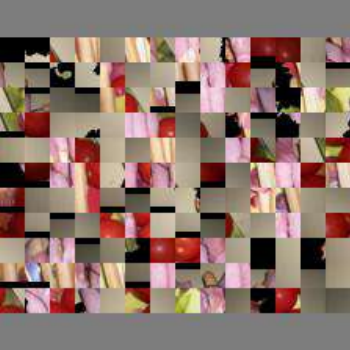} & \includegraphics[width=0.25\textwidth]{figs/Figure9-4.pdf}\\
\midrule
RveRNet  & \multirow{3}{*}{\underline{52.066}} & \multirow{3}{*}{\underline{16.373 (-35.693)}} & \multirow{3}{*}{\underline{36.427 (-15.639)}} & \multirow{3}{*}{\underline{8.624 (-43.442)}} \\
(ROI: MobileNetV2, & & & & \\
extra-ROI: MobileNetV2)& & & & \\
\midrule
RveRNet   & \multirow{3}{*}{55.76} & \multirow{3}{*}{28.395 (-27.365)} & \multirow{3}{*}{40.340 (-15.42)} & \multirow{3}{*}{15.523 (-40.237)} \\
(ROI: MobileNetV2, & & & & \\
extra-ROI: ViT-B/16)  & & & & \\
\midrule
RveRNet   & \multirow{3}{*}{55.181} & \multirow{3}{*}{35.539 (-19.642)} & \multirow{3}{*}{40.224 (-14.957)} & \multirow{3}{*}{20.826 (-34.355)} \\
(ROI: ViT-B/16, & & & & \\
extra-ROI: MobileNetV2)  & & & & \\
\midrule
RveRNet   & \multirow{3}{*}{55.322} & \multirow{3}{*}{35.603 (-19.719)} & \multirow{3}{*}{49.234 (-6.088)} & \multirow{3}{*}{25.885 (-29.437)} \\
(ROI: ViT-B/16, & & & & \\
extra-ROI: ViT-B/16)  & & & & \\
\midrule
RveRNet  & \multirow{3}{*}{55.902} & \multirow{3}{*}{41.472 (-14.43)} & \multirow{3}{*}{50.714 (\textbf{-5.188})} & \multirow{3}{*}{\textbf{30.107} (\textbf{-25.795})} \\
(ROI: DeiT-B, & & & & \\
extra-ROI: DeiT-B)  & & & & \\
\midrule
RveRNet   & \multirow{3}{*}{58.000} & \multirow{3}{*}{\textbf{44.575} (\textbf{-13.425})} & \multirow{3}{*}{48.526 (-9.474)} & \multirow{3}{*}{29.811 (-28.189)} \\
(ROI: DeiT-B, & & & & \\
extra-ROI: DeiT-B$\alambic$)  & & & & \\
\midrule
RveRNet   & \multirow{3}{*}{\textbf{61.076}} & \multirow{3}{*}{38.705 (-22.371)} & \multirow{3}{*}{\textbf{54.254} (-6.822)} & \multirow{3}{*}{28.331 (-32.745)} \\
(ROI: DeiT-$\alambic$, & & & & \\
extra-ROI: DeiT-B$\alambic$)  & & & & \\

\bottomrule
\end{tabular}
}
\end{table}

The top-1 accuracy dramatically dropped when permutation noise was introduced to the ROI modules’ inputs. However, when noise was introduced to the extra-ROI modules’ inputs and the ROI module was a transformer-based model, performance did not drop as much as when the ROI module was a CNN.

Moreover, performance dropped significantly less when the extra-ROI module input was permuted. This result indicates that overall classification performance was strongly correlated with ROI module choice. Therefore, the effect of noise in the extra-ROI module’s input can be diluted by the ROI module’s performance.

Due to the permutation invariance of the ROI transformer models ViT, DeiT, and DeiT$\alambic$, the proposed model was robust against permutation noise and its performance depended more on ROI module choice than extra-ROI module choice. These results indicate that DeiT was the transformer with the strongest encoded permutation invariance.
The permutation invariance of DeiT$\alambic$ was weaker than that of DeiT and ViT (Table 11). It can be assumed that DeiT’s strong permutation invariance gets weaker as a result of token-based distillation from a CNN teacher. However, DeiT$\alambic$ was better than other models at identifying features necessary for classification.

\subsection{Translation Invariance}

\begin{table}[t]
\caption{\textbf{Top-1 accuracy when the inputs of RveRNet were translocated resolution-wise.}
\label{tab:tab12}}
\smallskip
\centering
\scalebox{0.8}
{
\begin{tabular}{l|cccccccc}
\toprule
    &  \multicolumn{4}{c}{Top-1 accuracy (test set)} \\ \cline{2-5}
 & No data   &  \multicolumn{3}{c}{Translocation of extra-ROI module inputs} \\
& transformation   & dx: -30, dy: 63  & dx: 60, dy: -85 & dx: 37, dy: 139  \\ 

\midrule
ROI   &  \includegraphics[width=0.25\textwidth]{figs/Figure9-2.pdf}  & \includegraphics[width=0.25\textwidth]{figs/Figure9-2.pdf}  & \includegraphics[width=0.25\textwidth]{figs/Figure9-2.pdf}& \includegraphics[width=0.25\textwidth]{figs/Figure9-2.pdf} \\
\midrule
Extra-ROI & \includegraphics[width=0.25\textwidth]{figs/Figure9-3.pdf}   & \includegraphics[width=0.25\textwidth]{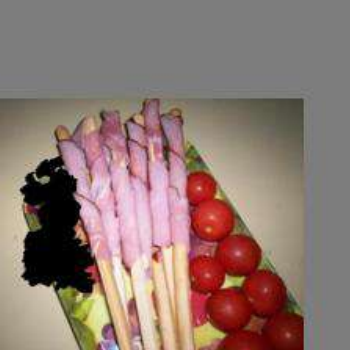} & \includegraphics[width=0.25\textwidth]{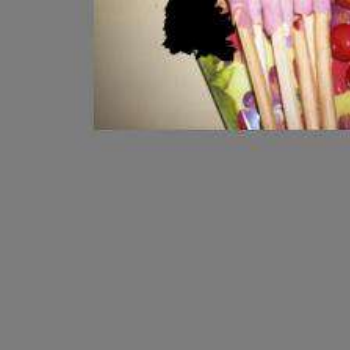} & \includegraphics[width=0.25\textwidth]{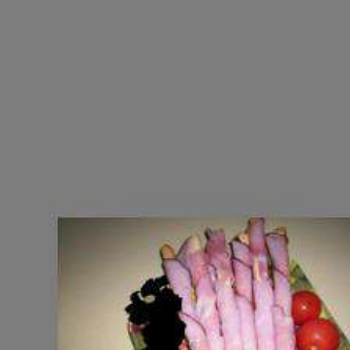}\\
\midrule
RveRNet  & \multirow{3}{*}{\underline{52.066}} & \multirow{3}{*}{\underline{51.216} (-0.850)} & \multirow{3}{*}{50.521 (-1.545)} & \multirow{3}{*}{\underline{48.745} (-3.321)} \\
(ROI: MobileNetV2, & & & & \\
extra-ROI: MobileNetV2)& & & & \\
\midrule
RveRNet   & \multirow{3}{*}{55.76} & \multirow{3}{*}{54.100 \underline{(-1.66)}} & \multirow{3}{*}{\underline{48.153 (-7.607)}} & \multirow{3}{*}{49.221 \underline{(-6.539)}} \\
(ROI: MobileNetV2, & & & & \\
extra-ROI: ViT-B/16)  & & & & \\
\midrule
RveRNet   & \multirow{3}{*}{55.181} & \multirow{3}{*}{54.988 (-0.193)} & \multirow{3}{*}{52.954 (-2.227)} & \multirow{3}{*}{53.134 (-2.047)} \\
(ROI: ViT-B/16, & & & & \\
extra-ROI: MobileNetV2)  & & & & \\
\midrule
RveRNet   & \multirow{3}{*}{55.322} & \multirow{3}{*}{54.563 (-0.758)} & \multirow{3}{*}{52.105 (-3.217)} & \multirow{3}{*}{52.401 (-2.921)} \\
(ROI: ViT-B/16, & & & & \\
extra-ROI: ViT-B/16)  & & & & \\
\midrule
RveRNet  & \multirow{3}{*}{55.902} & \multirow{3}{*}{57.858 (\textbf{+1.956})} & \multirow{3}{*}{55.258 (\textbf{-0.644})} & \multirow{3}{*}{55.103  (\textbf{-0.799})} \\
(ROI: DeiT-B, & & & & \\
extra-ROI: DeiT-B)  & & & & \\
\midrule
RveRNet   & \multirow{3}{*}{58.000} & \multirow{3}{*}{57.369 (-0.631)} & \multirow{3}{*}{53.006 (-4.994)} & \multirow{3}{*}{55.001 (-2.999)} \\
(ROI: DeiT-B, & & & & \\
extra-ROI: DeiT-B$\alambic$)  & & & & \\
\midrule
RveRNet   & \multirow{3}{*}{\textbf{61.076}} & \multirow{3}{*}{\textbf{59.982} (-1.094)} & \multirow{3}{*}{\textbf{57.511} (-3.565)} & \multirow{3}{*}{\textbf{58.231} (-2.845)} \\
(ROI: DeiT-B$\alambic$, & & & & \\
extra-ROI: DeiT-B$\alambic$)  & & & & \\

\bottomrule
\end{tabular}
}
\end{table}

We compared the performance of models and investigated how translation invariance, which is one of CNNs’ inductive biases, contributed to RveRNet’s classification performance when there was translocation noise in the extra-ROI module’s input.

CNNs were more robust to information loss accompanied by input translocation than all of the transformers except for DeiT (Table 12). When both modules of the RveRNet were DeiT, overall model performance decreased less than other combinations. This result showed that transformer-based architectures were robust against information loss caused by input translocation. However, RveRNet performed much worse when both modules were DeiT$\alambic$ than when both modules were DeiT. This result may be explained by the hypothesis that DeiT lost innate strength as the CNN’s knowledge was distilled to it.

\end{document}